\vspace{-0.5cm}
\documentclass[runningheads]{llncs}
\usepackage{graphicx}
\usepackage{arydshln}
\usepackage{multirow}
\usepackage{todonotes}
\begin{document}
\title{Predicting Soil pH by Using Nearest Fields}

\hyphenation{ontology}
%
%
\author{Quoc Hung Ngo \and
        Nhien-An Le-Khac \and
        Tahar Kechadi}
\authorrunning{Q.H. Ngo et al.}
%
\institute{University College Dublin, \\
           Belfield, Dublin 4, Ireland \\
\email{hung.ngo@ucdconnect.ie,\{an.lekhac,tahar.kechadi\}@ucd.ie}}

\maketitle              
\begin{abstract}
  In precision agriculture (PA),  soil sampling and testing operation is  prior to planting any
  new crop. It is an expensive operation since there are many soil characteristics to take
  into  account.   This  paper  gives  an  overview  of  soil  characteristics  and  their
  relationships with crop yield and soil  profiling. We propose an approach for predicting
  soil pH  based on nearest  neighbour fields.  It  implements spatial radius  queries and
  various regression  techniques in  data mining.   We use  soil dataset  containing about
  $4,000$ fields  profiles to evaluate  them and  analyse their robustness.  A comparative
  study  indicates that  LR, SVR,  and GBRT  techniques achieved  high accuracy,  with the
  \(R_2\) values of  about $0.718$ and \(MAE\) values of  $0.29$. The experimental results
  showed that the proposed approach is  very promising and can contribute significantly to
  PA.
\end{abstract}

\section{Introduction}
\label{sec:intro}
Precision agriculture  can be described  as an autonomous  process that collects  data and
presents  it to  analysis  systems to  mine  it. And  the application  of  data mining  to
agricultural data becomes highly important, as it is capable of mining huge collections of
data to look for new knowledge and, thus, improve the current practices.  In this context,
soil profile is  one of preconditions for making good  agronomic decisions. This practical
information  can be  obtained  by soil  sampling,  however,  it is  costly  and very  time
consuming. In  addition, it is often  not necessary to  conduct soil tests for  all fields
when the field conditions can be similar to the neighbourhood fields.

In general, the  use of data mining techniques  allows us to study a large  number of soil
profiles \cite{shangguan2013china} and monitor soil characteristics and other factors that
affect crop yield \cite{he2011soil} \cite{singh2012effects}.  These data mining techniques
have been  successfully used  to classify soil  data \cite{han2016smartphone},  to predict
soil map  \cite{da2017data} and soil  salinity \cite{wang2019comparison}. However,  to the
best of our knowledge, there is no study on predicting soil characteristics for new fields
with only their  locations and some other  features. Prediction of soil  features based on
nearest fields  not only  supports to  fill omitting  values for  soil profiling  but also
reduces cost for soil sampling.

In  this  paper, we  propose  a  solution to  generate  features  for new  fields  without
sampling. This is  of great help, mainly  when some data values were  missing during their
collection. We  also propose a  data mining  approach to predict  soil pH values  based on
neighbourhood field values.  Finally, we test  and evaluate our approach experimentally on
real data collected from  about $4,000$ soil profiles. The next  section gives an overview
of  soil  properties and  reviews  several  soil studies  that  are  related to  precision
agriculture.

\section{Related Work}
\label{sec:RW}
The most important soil characteristics can be divided into three categories: composition,
physical  and  chemical  characteristics  \cite{osman2013soils}. In  addition,  there  are
several features, which  relate to soil fertiliser and biological  properties, such as CEC
(Cation  exchange  capacity),   SOC  (soil  organic  carbon),  and   EC  (Soil  electrical
conductivity).    In  fact,   soil   profiles  mainly   include   physical  and   chemical
characteristics  (such as  pH, N,  P, K,  etc), SOC,  or SEC  \cite{bishop2001comparison},
\cite{shangguan2013china},\cite{singh2012effects}.    These   soil  characteristics   were
already represented using the AgriOnto  ontology \cite{ngo2018ontology}.  Many studies can
be  found  in the  literature  on  building soil  profiles  or  datasets, monitoring  soil
characteristics that affect  crop yield. Many of those studies  use data mining techniques
on soil characteristics to predict crop yield and other measures or objectives.

Wei Shangguan et al. \cite{shangguan2013china} built  a China soil dataset of $8,979$ soil
profiles  for land  surface  modelling. The  data  set includes  $28$  attributes for  $8$
vertical layers (from  $0$ to $2.296$m) which were collected  from $2,444$ counties, $312$
national  farms  and  $44$  forest  farms.   In   an  other  soil  study,  P.K.  Singh  et
al. \cite{singh2012effects}  monitored pH, EC,  CEC, and chemical characteristics  of soil
samples during  and after crop harvesting  to evaluate the  effect of waste water  on soil
properties, crop yield and the environment.

In  P. Han  et  al.  \cite{han2016smartphone}, soil  colour  characteristics  are used  to
classify  $10$ soil  types. They  used soils  layers  in a  depth of  $40-80$cm below  the
surface.  Their classifier is based on  RGB signals and principal component analysis (PCA)
to classify  the data.  The  experimental findings have been  obtained and evaluated  on a
data set of $50$ soil samples per soil type ($500$ samples in total).

For  the  prediction  of  soil  characteristics,  authors  in  \cite{bishop2001comparison}
compared $3$ prediction methods  for mapping CEC. Their study was carried  out on a $74$ha
field  in   Australia  for   a  duration   of  $2$  years   ($1996$  sorghum   and  $1997$
wheat). \cite{wang2019comparison} predicted soil salinity in three geographically distinct
areas in  China. They  compared five regression  algorithms based on  $21$ data  sets with
$189$ soil samples to predict soil salinity.  In their experiments, random forest (RF) and
stochastic gradient  treeboost (SGT) achieved the  highest accuracy with \(R^2\)  score of
$0.63$.   However, the  scores  of RF  and  SGT predictions  are not  stable  by time  and
locations.  The most  wide range of soil chemical and  physical characteristics prediction
was  conducted by  M.J.   Aitkenhead et  al.   \cite{aitkenhead2013prediction}. They  used
artificial neural  networks (ANN)  and the soil  color (RGB values)  to predict  $44$ soil
parameters  including  chemical, physical  characteristics  and  soil texture.  They  also
demonstrated that several soil parameters can  be predicted accurately (with \(R^2\) \(>\)
$0.5$).

In summary,  several studies on the  use of data  mining techniques to predict  other soil
characteristics  of  each   soil  profile  or  predict  other  factors   related  to  soil
characteristics have  been presented  in the  literature.  To the  best of  our knowledge,
there   is  not   any  study   on  predicting   new  soil   profiles  without   main  soil
characteristics.  Moreover, previous  studies on  soil  profile prediction  based on  soil
characteristics constitute a solid foundation for us to carry on this work.

\section{Predicting Soil pH}
\label{sec:PSpH}
\subsection{Soil Dataset}
\label{sec:SD}
The soil  dataset includes  soil sampling of  $3,809$ fields, which  are extracted  from a
large raw agriculture dataset of the CONSUS project. The soil datasets were collected from
a widely distributed agriculture area of the  UK. These fields grow many different plants,
but the collected datasets were mainly focus on crops, fruits, vegetables, and grass.
Each record  in the dataset  corresponds to one  field, which includes  field information,
location information  (longitude, latitude), chemical  features (pH,  P, K, Mg),  and soil
texture (sand,  clay, and silt percentage).   According to \cite{pietri2008relationships},
soil pH is  the most important attribute.  The values of this attitude are between $0$ and  
$14$, but,  they are mainly  from $5$  to $8.5$ for  cultivated fields  in our dataset.

\subsection{Features based on Nearest Fields}

The  number of  fields, which  have nearest  fields within  the radius  of $250$m,  is the
highest     and  most fields have neighbours within the radius of $2,000$m (3,760 of 3,809 
fields,   as shown in Table \ref{tableFieldNeighbour}). But, there are several fields that 
only  have nearest fields in the radius of $10,000$m ($10$km). There are about $50$ fields 
without neighbours within the radius of $2,000$m.
\begin{table}[h]
  \caption{Validate soil feature by nearest fields}
  \label{tableFieldNeighbour}
  \setlength{\tabcolsep}{4pt}
  \begin{center}
    \begin{tabular}{|r|r|r|r|r|}
      \hline
      \textbf{Radius} & \textbf{Fields have} 
      & \textbf{Number of}  & \textbf{Distance} & \textbf{Average of}\\
      \textbf{(m)}    & \textbf{neighbours}  & \textbf{Neighbours} 
                            & \textbf{(m)} & \textbf{max-min(pH) }\\
      \hline
        100 &    25 &  1.12 &  78.2  & 0.03 \\
        200 &   756 &  1.28 & 147.42 & 0.09 \\
        300 & 2,102 &  1.67 & 185.22 & 0.19 \\
        400 & 2,945 &  2.27 & 210.35 & 0.31 \\
        500 & 3,295 &  3.01 & 232.57 & 0.44 \\
        750 & 3,594 &  5.07 & 296.29 & 0.67 \\
      1,000 & 3,672 &  7.11 & 367.19 & 0.83 \\
      1,500 & 3,733 & 10.65 & 505.93 & 1.04 \\
      2,000 & 3,760 & 13.66 & 635.17 & 1.16 \\
      \hline
    \end{tabular}
  \end{center}
  \vspace{-2em}
\end{table}

Our approach is based  on field's location. For each field in the  dataset, we can get the
nearest fields that are  within a radius of a given field (based  on spatial queries). The
radius  (in meters)  is the  maximum  allowed distance  between  the given  field and  the
returned list  of nearby fields.  In our  experience, the radius  is in the  range between
$200-2,000$m.

To  predict the  pH attribute  of data  object  \(y\) (or  field \(y\)),  we estimate  the
average, maximum,  and minimum pH values  based on the pH  values of the returned  list of
nearby fields of  \(y\) and the distance \(Dist\)  between the centre of the  list and the
location of \(y\).

\[pH_{avg}(y,r_i)=\frac{\sum_{j=1}^{k}pH(x_j)}{k}; Dist(y,r_i) = distance(y, x_{centre});\]
\[pH_{min}(y,r_i)=min(pH(x_j));pH_{max}(y,r_i)=max(pH(x_j));\]
where \(k\) is  the number of neighbours  in the radius of \(r_i\)  (e.g.  $200$m, $300$m,
$400$m, and  $500$m), \(x_j\) is  the neighbour field in  this region (\  j=\{1..k\}), and
\(x_{centre}\) is the centre of \(k\) neighbours for each radius \(r_i\)(m).

\subsection{Data Mining Techniques for Prediction}

There are many data mining techniques used for soil classification and prediction.  In our
study, we  propose to  use common  data regression  techniques to  predict soil  pH. These
techniques   include   Linear   regression   (LR),   Support   Vector   regression   (SVR)
\cite{basak2007support}, Decision Tree  Regression (DTR) \cite{waheed2006measuring}, Least
Absolute Shrinkage and Selection  Operator (LASSO) \cite{tibshirani1996regression}, Random
Forests  (RF)  \cite{breiman2001random},  and  Gradient Boosting  Regression  Tree  (GBRT)
\cite{breiman2001random}.    In    our   experiments,   we   use    Scikit-learn   toolkit
(https://scikit-learn.org/) to deploy and evaluate these techniques.

\section{Experimental Results}
\label{sec:ER}

In our experiments,  the comparative evaluation of  the prediction models is  based on the
coefficient  of determination  (\(R^2\))  and  the mean  absolute  error  (MAE). The  best
possible coefficient score \(R^2\) is $1.0$ and  the worst is $0.0$. A constant model that
always predicts  the expected  value of y,  disregarding the input  features, would  get a
\(R^2\) score of 0.0.

In the first experiment, we apply six  regression techniques (LR, SVR, LASSO, DTR, RF, and
GBRT) on a  part or the whole  dataset depending on the evaluated  features.  For example,
when evaluating  a group of features  related to the  radius of $200$m, only  $756$ fields
have    neighbour    fields,    therefore    the   size    of    data    for    evaluating
\textit{CropName+Min/Max/Avg200} features is ($756, 4$) (Table \ref{tableIndiResult}). The
obtained results for Soil  pH prediction were very low with owned  field features (1st row
of Table \ref{tableIndiResult}). The results improved significantly when adding average pH
features. We achieved  high results with \textit{CropName+Min/Max/Avg400} features.

\begin{table*}[h]
    \caption{Result of Soil pH regression based on radius-based features \newline
             \(llcn^*\): Long/Lat/CropName; \(r200^{**}\): CropName+Min/Max/Avg200}
    \label{tableIndiResult}
    \setlength{\tabcolsep}{2pt}
    \begin{center}
        \begin{tabular}{crrrrrrrrrrrr}
        \hline
            \multirow{2}*{\textbf{Attr.}} & 
            \multicolumn{2}{c}{\textbf{LR}} & \multicolumn{2}{c}{\textbf{SVR}} &
            \multicolumn{2}{c}{\textbf{LASSO}} & \multicolumn{2}{c}{\textbf{DTR}} &
            \multicolumn{2}{c}{\textbf{RF}} & \multicolumn{2}{c}{\textbf{GBRT}} \\
            & \(R^2\) & MAE & \(R^2\) & MAE & \(R^2\) & MAE & \(R^2\) & MAE & 
                \(R^2\) & MAE & \(R^2\) & MAE \\
        \hline
            \(llcn^*\) & 0.084 & 0.56 & 0.163 & 0.52 & 
                -0.004 & 0.61 & 0.46 & 0.36 & 0.162 & 0.52 & 0.536 & 0.35 \\
            \(r200^{**}\) & 0.681 & 0.33 & 0.688 & 0.33 & 
                -0.001 & 0.62 & 0.47 & 0.41 & 0.666 & 0.35 & 0.66 & 0.33 \\
            r300 &  0.695 & 0.29 & 0.698 & 0.29 & 
                -0.004 & 0.6 & 0.427 & 0.39 & 0.66 & 0.31 & 0.683 & 0.29 \\
            r400 &  \textbf{0.718} & \textbf{0.29} & \textbf{0.713} & \textbf{0.28} & 
                -0.002 & 0.63 & 0.503 & 0.39 & 0.68 & 0.31 & \textbf{0.703} & \textbf{0.29} \\
            r500 &  0.671 & 0.31 & 0.666 & 0.3 & 
                -0.001 & 0.61 & 0.411 & 0.4 & 0.654 & 0.32 & 0.651 & 0.31 \\
            r750 &  0.633 & 0.28 & 0.632 & 0.27 & 
                -0.0 & 0.62 & 0.386 & 0.4 & 0.645 & 0.3 & 0.628 & 0.28 \\
            r1000 &  0.66 & 0.31 & 0.663 & 0.31 & 
                -0.001 & 0.61 & 0.452 & 0.4 & 0.617 & 0.32 & 0.669 & 0.3 \\
            r1500 &  0.653 & 0.3 & 0.647 & 0.29 & 
                -0.001 & 0.62 & 0.454 & 0.36 & 0.62 & 0.32 & 0.656 & 0.29 \\
            r2000 &  0.623 & 0.33 & 0.608 & 0.32 & 
                -0.002 & 0.62 & 0.452 & 0.38 & 0.596 & 0.35 & 0.658 & 0.32 \\
        \hline
        \end{tabular}
    \end{center}
\end{table*}

In another  experiment, we  evaluated the  contribution of  features to  prediction.  Only
three regression techniques have returned high scores; these are LR, SVR and GBRT. We have
also evaluated the \textit{CropType} feature, which  represents a mapping of the crop name
to a  crop type list (including  Crops, Vegetables, Fruits,  and Grass) by using  lists of
concepts and  instances from the  AgriOnto ontology \cite{ngo2018ontology}.   Although the
\textit{CropName} feature  contains over $40$  different crop names,  it is mapped  to the
\textit{CropType} feature with  four crop types.   The results  are approximately the same 
for both experiments (3rd, 4th row of Table \ref{tableFeatureResult}).

\begin{table}[h]
    \caption{\(R^2\) score of Soil pH regression based on individual features}
    \label{tableFeatureResult}
    \setlength{\tabcolsep}{6pt}
    \begin{center}
        \begin{tabular}{ccrrr}
        \hline
            \textbf{Feature} & \textbf{Size} & \textbf{LR} & \textbf{SVR} & 
            \textbf{GBRT} \\
        \hline
            Long/Lat/CropName & (2945, 3) & 0.086 & 0.17 & 0.548 \\
            Long/Lat/CropName+Avg400 & (2,945, 4) & 0.717 & 0.715 & 0.716 \\
            Nb/Dist/Avg400+CropName & (2,945, 4) & 0.717 & 0.714 & 0.7 \\
            Nb/Dist/Avg400+CropType & (2,945, 4) & 0.718 & 0.714 & 0.696 \\
            Nb/Dist/Max/Min/Avg400 & (2,945, 5) & 0.718 & 0.709 & 0.697 \\
            + CropName & (2,945, 6) & 0.718 & 0.708 & 0.696 \\
            + CropName, CropType & (2,945, 7) & 0.718 & 0.709 & 0.696 \\
            \hline
        \end{tabular}
    \end{center}
    \vspace{-2.5em}
\end{table}

In  the next  experiments, we  extended  the number  of  features to  include more  radius
values. As shown in Table \ref{tableCombineResult},  it calculates the average pH value of
neighbours  in the  radius  ranging  from $200$m  to  $2,000$m.     The same    algorithms
achieved their highest scores at  the radius values $400$m and $500$m. 

\begin{table}[htb]
    \caption{\(R^2\) score of Soil pH regression based on combined features}
    \label{tableCombineResult}
    \setlength{\tabcolsep}{8pt}
    \begin{center}
        \begin{tabular}{ccrrr}
        \hline
            \textbf{Feature} & \textbf{Size} & \textbf{LR} & \textbf{SVR} & \textbf{GBRT} \\
        \hline
            Long/Lat/CropName & (756, 3) & 0.122 & 0.232 & 0.545 \\
            + Nb/Dist/Avg200 & (756, 6) & 0.686 & 0.667 & 0.656 \\
            + Nb/Dist/Avg300 & (756, 9) & 0.699 & 0.676 & 0.674 \\
            + Nb/Dist/Avg400 & (756, 12) & 0.715 & 0.67 & 0.684 \\
            + Nb/Dist/Avg500 & (756, 15) & 0.711 & 0.638 & 0.692 \\
            + Nb/Dist/Avg750 & (756, 18) & 0.709 & 0.597 & 0.702 \\
            + Nb/Dist/Avg1000 & (756, 21) & 0.707 & 0.58 & 0.696 \\
            + Nb/Dist/Avg1500 & (756, 24) & 0.704 & 0.604 & 0.703 \\
            + Nb/Dist/Avg2000 & (756, 27) & 0.702 & 0.493 & 0.697 \\
        \hline
        \end{tabular}
    \end{center}
\end{table}

\section{Conclusion and Future Work}
\label{sec:CFW}

We presented a  short study on soil  properties and how to  construct soil profiles which 
can be sued in crop  yield management. We proposed an approach  to predict  soil pH based 
on the average  pH values of the nearest neighbour fields. This can be applied to predict 
other characteristics of the soil profile if these characteristics were missing.
With large  soil dataset,  our approach    based  only on  neighbour fields  has a  great
potential   not only    for pH prediction  but also to  predict other soil features. As a  
result, we plan to  extend our model and  perform more experiences to  predict other soil 
characteristics.  Moreover,  the weather data or crop yield  are also highly valuable  to 
add into prediction models.

\paragraph{\textbf{Acknowledgment}} This work  is part of CONSUS and  is supported by the 
the SFI Strategic Partnerships Programme    (16/SPP/3296)   and    is co-funded by Origin 
Enterprises Plc.

\end{document}